\pdfoutput=1
\documentclass[11pt]{article}

\usepackage[]{naacl2021}

\usepackage{times}
\usepackage{latexsym}

\usepackage[T1]{fontenc}
\usepackage[utf8]{inputenc}

\usepackage{microtype}
\usepackage{booktabs}
\usepackage{array}
\usepackage{makecell}
\usepackage{adjustbox}
\usepackage{subcaption}
\usepackage{enumitem}
\setlist{nosep}
\newcommand{\ignore}[1]{}

\newcommand{\hancomment}[1]{}

\newcommand{\psrcomment}[1]{}

\newcommand{\joecomment}[1]{}

\newcommand{\dougcomment}[1]{}

\newcommand{\cshivadecomment}[1]{}

\title{Towards Clinical Encounter Summarization: \\
Learning to Compose Discharge Summaries from Prior Notes}

\author{Han-Chin Shing\textsuperscript{$\dagger$}, 
  Chaitanya Shivade\textsuperscript{$\ddagger$},
  Nima Pourdamghani\textsuperscript{$\ddagger$},
  Feng Nan\textsuperscript{$\ddagger$}, \\
  {\bf Philip Resnik\textsuperscript{$\dagger$},
  Douglas Oard\textsuperscript{$\dagger$},
  Parminder Bhatia\textsuperscript{$\ddagger$}} \\
  \textsuperscript{$\dagger$}University of Maryland, 
  \textsuperscript{$\ddagger$}Amazon Web Service AI \\
  \texttt{\{shing,resnik,oard\}@umd.edu} \\
  \texttt{\{shivadc,nimpourd,nanfen,parmib\}@amazon.com}
  }

\begin{document}
\maketitle
\begin{abstract}
% Background

%\hancomment{TODO: rewrite abstract when close to finish}

% Physicians spent double the amount of time on documentation compared to patient visit. This motivates towards automated tools that can extract and summarize the relevant information. To reduce physician burnout, our work introduces the task of generating discharge summaries for a clinical encounter.
The records of a clinical encounter can be extensive and complex, thus placing a premium on tools that can extract and summarize relevant information. 
This paper introduces the task of generating discharge summaries for a clinical encounter.
Summaries in this setting need to be faithful, traceable, and scale to multiple long documents, motivating the use of extract-then-abstract summarization cascades.
We introduce two new measures, faithfulness and hallucination rate for evaluation in this task, which complement existing measures for fluency and informativeness.
Results across seven medical sections and five models show that a summarization architecture that supports traceability yields promising results, and that a sentence-rewriting approach performs consistently on the measure used for faithfulness (faithfulness-adjusted $F_3$) over a diverse range of generated sections.
\end{abstract}

\section{Introduction}
\label{sec:intro}

\begin{figure*}
    \centering
    \begin{adjustbox}{width=0.8\textwidth}
    \includegraphics{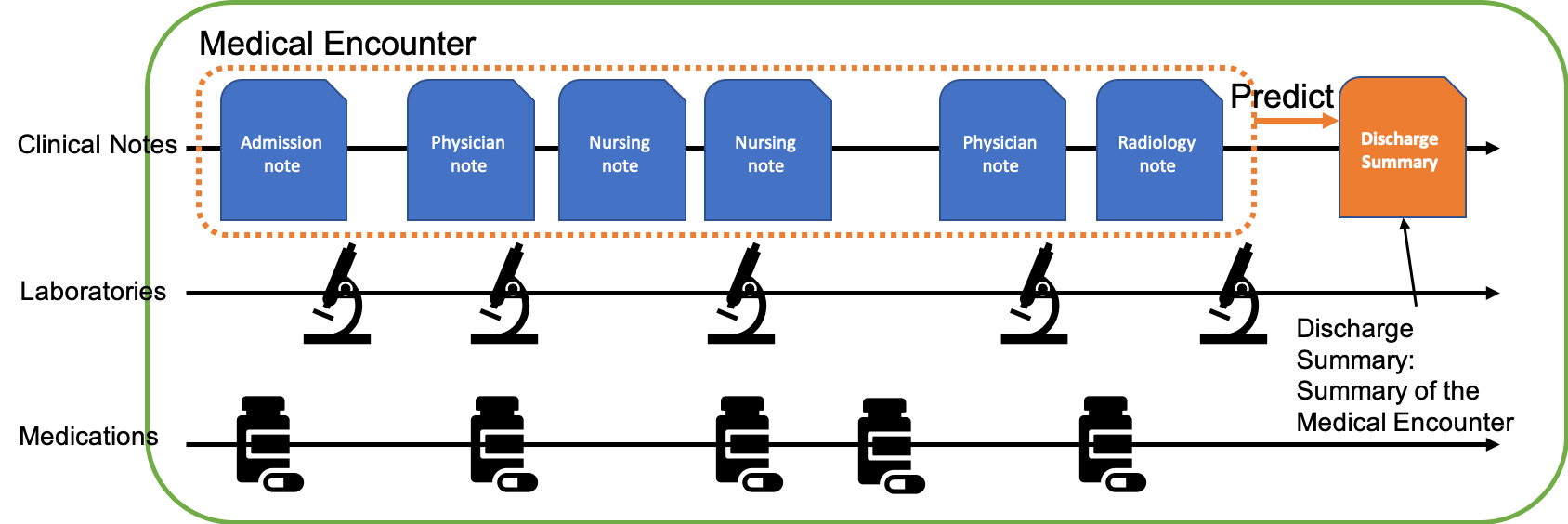}
    \end{adjustbox}
    \caption{A medical encounter is an interaction between a patient and a healthcare provider.}
    \label{fig:medical_encounter}
    \vspace{-1em}
\end{figure*}

\ignore{The abundance of clinical information has a clear potential in assisting clinicians in making informed clinical decisions. This potential, however, is limited by the fact that clinicians already don't have enough time to handle both patients and their clinical data~\citep{weiner2006influence,sinsky2016allocation}. The need to spend more time to document and interpret the patients' data without an effective way to process and manipulate them can even contribute to the worsening physician burnout crisis~\citep{tawfik2018physician,West2018Physician}.}

%These clinical notes also are the medium that serves as a communication channel between different, and sometime future, clinicians.

Clinical notes in the electronic health record (EHR) are used to document the patient's progress and interaction with clinical professionals. 
These notes contain rich and diverse information, including but not limited to admission notes, nursing notes, radiology notes, and physician notes. 
The information the clinicians need, however, is often buried in the sheer amount of text, as the number of clinical notes in an encounter can be in the hundreds. Finding the information can be time-consuming; time that is already in short supply for the clinicians to attend to the patients~\citep{weiner2006influence,sinsky2016allocation}, and can even contribute to the worsening physician burnout crisis~\citep{tawfik2018physician,West2018Physician}.

Summarization has the potential to help clinicians make sense of these clinical notes. 
In this paper, we aim to make progress toward summarizing one of the most common information sources clinicians interact with --- the patient's clinical encounter.
A clinical encounter (Figure~\ref{fig:medical_encounter}) documents an interaction between a patient and a healthcare provider (e.g., a visit to the hospital), including structured and unstructured data. Our work focuses on the unstructured clinical notes. 

A natural target for summarization is the \textit{discharge summary}: a specialized clinical note meant to be a summary of the clinical encounter, typically written at the time of patient discharge. 
Each section (e.g., past medical history, brief hospital course, medications on admission) in the discharge summary represents a different aspect of the encounter. 
By building a system to extract and compose these medical sections from prior clinical notes in the same encounter, we can summarize the information in a format clinicians are already trained to read and understand.

There are significant challenges ahead, however. 
In this work, we identify three main challenges of summarizing a clinical encounter: (1) an \emph{evidence-based fallback} that allows traceable inspection, (2) the \emph{faithfulness} of the summary, and (3) the \emph{long text} in a clinical encounter. 
We believe that all three challenges need to be properly addressed before a discussion about deployment can happen. 
Thus, this work focuses on measuring and understanding how existing state-of-the-art summarization systems perform on these challenges. 
Additionally, we propose an extractive-abstractive summarization pipeline that directly addresses the \emph{evidence-based fallback} challenge and the \emph{long text} challenge. 
For the third challenge, \emph{faithfulness}, we introduce an evaluation measure that uses a medical NER system, inspired by recent work on faithfulness in summarization~\citep{maynez2020faithfulness,zhang-etal-2020-optimizing}.

\vspace{-0.5em}
\paragraph{Contributions} 
%We create the task of discharge summary composition from prior documents. We identify three challenges from the context of summarization of many, long clinical documents: (1) faithfulness, (2) evidence, and (3) scalability. We address these challenges by evaluating with medical NER-based scores, and design an extractive-abstractive pipeline for multi-document summarization. By deriving from a public database (MIMIC III), the task can potentially serve as a benchmark for clinical multi-document summarization.

\begin{itemize}[leftmargin=0.9em,itemsep=-\parsep]
    \item We identify three challenges for summarizing clinical encounters: (1) faithfulness, (2) evidence, and (3) long text.
    \item We introduce the task of discharge summary composition from prior clinical notes.
    \item We evaluate our proposed extractive-abstractive pipeline for multi-document summarization with medical NER-based scores and ROUGE across seven discharge summary sections.
    \item We create a collection derived from a public database (MIMIC III); a potential benchmark for clinical multi-document summarization.
\end{itemize}

% What is the problem? Why is it important? What makes it difficult? What is the falsifiable hypothesis?

% The problem:

% Summarizing encounter notes for clinicians. 

% Importance:

% Alleviate the documentation and reading burden.

% Why is it a toward paper:

% The danger of model hallucination, its effect on people's lives.

% Difficulties:

% Faithfulness, Evidence-based fallback, Long text

% Falsifiable hypothesis:

% An extractie-abstractive pipeline can help address these issue. A faithfulness measure is designed to measure the parts of this pipeline.

\section{Evidence, Faithfulness, and Long Text}
\vspace{-0.5em}

%For a summarization system that can potentially be used to support clinical decision making, 

In this section, we identify the three main challenges in discharge summary composition.

\paragraph{\textbf{Evidence.}} 
A summary should be displayed with means for the clinician to inspect and understand where the \ignore{basis of summarization comes}information come from. In this respect, extractive summarization has a clear advantage over abstractive summarization, as the source of extracted content can be easily traced and displayed in context. However, abstractive summarization does benefit from a more fluent generation and the potential to function as a writing aid to alleviate the clinicians' documentation burden. The challenge lies in how to design the system such that \emph{evidence} can be traced.

\paragraph{\textbf{Faithfulness.}} Like any model supporting clinical decision making, measuring and understanding the faithfulness of the model output is important. As abstractive summarization systems are evaluated by their ability to generate fluent output, faithfulness can be a challenge to these models. Addressing this problem is an active area of research~\citep{maynez2020faithfulness,zhang-etal-2020-optimizing}.

\paragraph{\textbf{Long text.}} Summarizing an encounter (a sequence of documents), the quantity of text available can easily exceed the memory limit of the model. This memory limitation is especially challenging for modern transformer-based architectures that typically require large GPU-memory. Tokens that do not fit in memory can contain relevant clinical information for summarization. Attempting to train an abstractive model to generate a summary without the source information available can encourage the model to hallucinate at test time; a dangerous outcome in the context of clinical summarization.

\vspace{-0.5em}

\section{Extract and then Abstract}

% Benefit of extractive-abstractive pipeline (with more weight on the extractor): Fallbacks, long text, and hallucination.
\label{sec:address}

\begin{figure*}[ht]
    \centering
    \adjustbox{width=0.7\textwidth}{
    \includegraphics{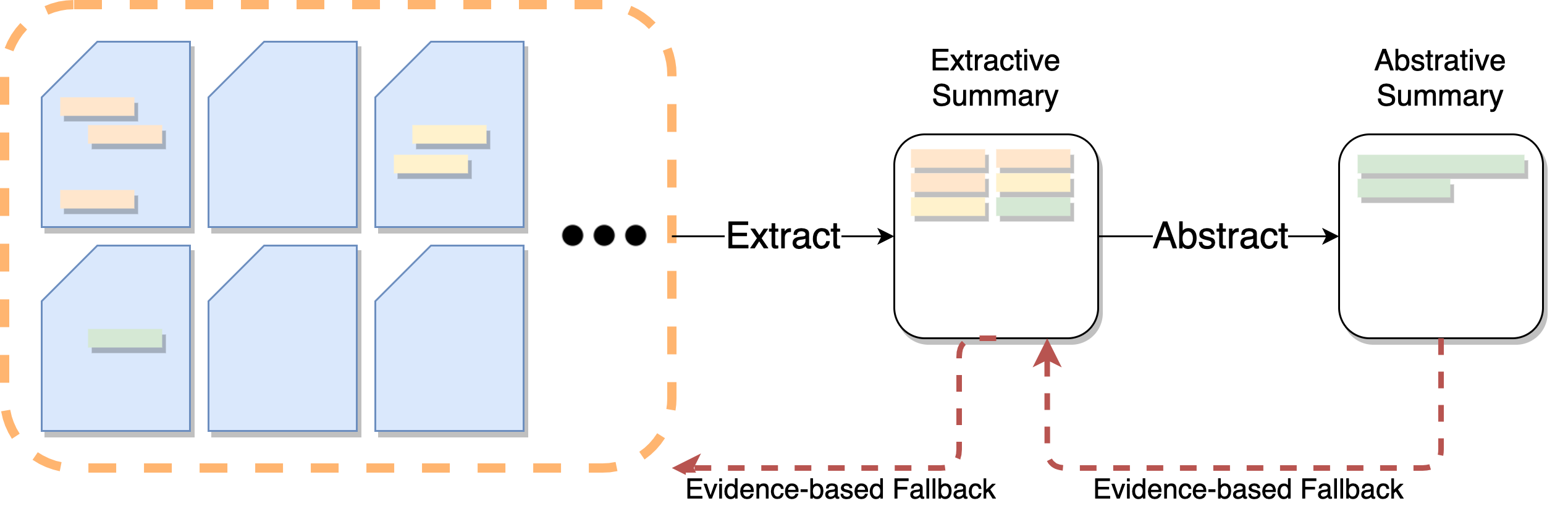}
    }
    \caption{\hancomment{TODO: update figure with better graphic} An extractive-abstractive summarization pipeline. The recall-oriented extractor extract relevant sentences from prior documents, the abstractor smooths out irrelevant or duplicated information.}
    \label{fig:pipeline}
    \vspace{-1em}
\end{figure*}

\vspace{-0.5em}

%We argue that this re-ranker can be viewed as a type of extractive model. This inherently two-stage process hints at an extractive-abstractive pipeline.

These challenges are common in summarization. In particular, one of the main challenges in multi-document abstractive summarization is to summarize a large number of documents. While significant progress has been made to scale the abstractive models~\citep{Beltagy2020Longformer,zaheer2020big}, recent work still involves using an extractive model (e.g., tf-idf based cosine similarity~\citep{liu2018generating}, logistic regression~\cite{liu2019hierarchical}) to limit the number of paragraphs before abstraction. 

Here we proposed a similar extractive-abstractive pipeline. However, what is different in a clinical context is that we wish to place more weight on the extractor than rely on the abstractor to summarize a large amount of text. This decision is motivated by the fact that extractive models are inherently better at being faithful to the source, as they do not introduce novel information. This characteristic makes them ideal candidates for clinical summarization.

Our proposed extractor-abstractor pipeline involves two stages (Figure~\ref{fig:pipeline}). The first stage functions as a recall-oriented extractive summarization system to extract relevant sentences from prior documents. The extracted sentences are then passed through post-processing steps that remove duplicated sentences and arrange them to form an extractive summary.
The second stage is an abstractive summarization system that aims to take the extractive summary from the previous step and smooths out irrelevant or duplicated information.
We describe the details of implementations and how to scale this pipeline to very long text in Section~\ref{sec:model}.

Another advantage of this pipeline is that it provides a clear path of \emph{evidence-based fallbacks} (Figure~\ref{fig:pipeline}). Notably, both the extractor and the abstractor in the extractor-abstractor pipeline are capable of producing full summaries.
Clinicians can reference the extractive summary if they find the abstractive summary problematic or if the abstractor model has low confidence. The extractive summary also has another level of fallback. The extracted sentences came from the source documents, so we can also display the extracted sentences in context or even use the extractor as a highlighter.

\section{Measuring Faithfulness}
\label{sec:measure}

\begin{figure}[t]
    \centering
    \adjustbox{width=0.5\columnwidth}{
    \includegraphics{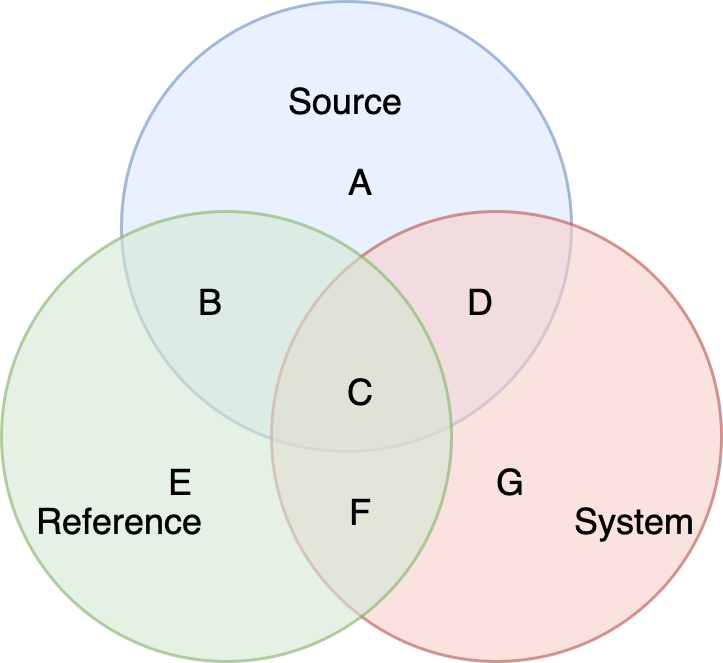}
    }
    \caption{Relationship between source documents, reference summary, and system-generated summary.}
    \label{fig:venn_diagram}
    \vspace{-1em}
\end{figure}

Following prior work, we report ROUGE-n ($n=\{1,2\}$) to measure n-gram overlap as a proxy for informativeness, and ROUGE -L (longest common subsequence, with possible gaps) as a proxy for fluency~\citep{lin2003automatic, maynez2020faithfulness}. However, as~\citet{schluter2017limits} and~\citet{cohan2016revisiting} have argued, ROUGE alone is insufficient and possibly misleading for measuring informativeness, specifically when it comes to faithfulness and factualness.

In a summarization setting, a faithful summary refers to a summary that does not contain information from outside of the source. On the other hand, a factual summary allows information not presented in the source, as long as the information is factually correct. 
%While the summarization community has not yet reached a consensus on whether faithfulness or factualness should be the end target~\citep{maynez2020faithfulness}, 
In the setting of clinical summarization, we argue that faithfulness is far more important. 
%The role of summarization in a clinical setting should be assisting the decision making process of human, not being the decision maker. 
Novel information appearing in a summary that has no support from the source, whether factual or not, can affect the transparency of the model.

A downside of this definition of faithfulness, however, is that it does not take reference summaries into account. Any extracted sentences (e.g., the first three sentences) from the source are always faithful by definition. Such extraction, however, might not be a summary relevant to this task. Figure~\ref{fig:venn_diagram} helps us illustrate the relationship between source documents, reference summary, and system-generated summary using a Venn diagram.\footnote{Here we are showing a single reference summary, but in reality, the reference summary available is just one possible manifestation of all possible, potentially equally valid summaries~\citep{nenkova2004evaluating}. Our discussion can be extended to multiple reference summaries by treating each one independently in the calculations and averaging them to report the final scores.} 

A desirable summary, especially in a clinical setting, is faithful to the source and relevant as measured by the reference summary. In Figure~\ref{fig:venn_diagram}, this region corresponds to $B + C$, the ideal set of information a clinical summarization system should target. Based on this observation, we define \emph{Faithfulness-adjusted Precision} as $\frac{C}{System}$ and \emph{Faithfulness-adjusted Recall} as $\frac{C}{B+C}$. Intuitively, faithfulness-adjusted precision measures how much information in the system-generated summary is both relevant and faithful. Similarly, faithfulness-adjusted recall measures the amount of faithful and relevant information that has been included by the system. In a clinical setting, recall is often more important than precision; it is better to over-extract and have clinicians ignore or remove the irrelevant content than have missing content. While our extractive-abstractive pipeline provides a series of \emph{fallbacks} that allows clinicians to inspect what could be missing by looking at the context of the extracted sentences, under-extraction can still happen. We therefore report a recall-oriented measure to combine the two above measures: \emph{Faithfulness-adjusted F\textsubscript{$\beta$}}, where we set $\beta = 3$. In this setting, faithfulness-adjusted recall is three times more important than faithfulness-adjusted precision~\citep{rijsbergen1979information}.\footnote{We plan to explore the values of $\beta$ in consultation with clinicians in future work.}

Hallucination is perhaps the leading concern of applying abstractive summarization system in a clinical setting. If one defines hallucination as a system generating content that is not faithful to the source, we can identify hallucination as the region $F + G$. $G$, the information that is not present in neither the source nor the reference, is particularly problematic. We therefore measure \emph{Incorrect Hallucination Rate} as $\frac{G}{System}$.

However, an important underlying assumption of these measures is that the regions in Figure~\ref{fig:venn_diagram} are quantifiable. While there are many ways to approximate these regions, as a starting point, we use the medical named entity recognition (NER) system in SciSpacy~\citep{neumann-etal-2019-scispacy}. The SciSpacy NER matches any spans in the text which might be an entity in UMLS, a large biomedical database, and transforms the text into a set of medical entities. The cardinalities of the sets and their overlaps can then be used to calculate the above measures.

\section{Related Work}

%Clinical abstractive summarization is not limited to radiology reports. 

\paragraph{Clinical Summarization.} Most literature on clinical summarization focuses on extractive summarization, due to the risk involved in a clinical application~\citep{demner2006answer,FEBLOWITZ2011688,liang2019novel,MOEN201625}. \hancomment{TODO: more details needed, will need to expand on this} 
For abstractive summarization, summarization of radiology reports has been a topic of interest in NLP research recently. \newcite{Zhang2018} show promising results generating assessment section of a chest x-ray radiology report from the findings and background section. \newcite{MacAvaney2019} improved this model through the incorporation of domain-specific ontologies. However, such generated reports may not be clinically sound, and the models generate sentences inconsistent with the patient's background. Therefore, in subsequent work \cite{zhang-etal-2020-optimizing} added a reinforcement learning based fact-checking mechanism to generate a clinically consistent assessment.
\newcite{Lee2018} explores the generation of the \textit{Chief Complaint} of emergency department cases from age group, gender, and discharge diagnosis code. \newcite{Ive2020} follow a closely related approach of extracting keyphrases from mental health records to generate synthetic notes. They further evaluate the quality of generated synthetic data for downstream tasks. Work from \newcite{Lee2018} generates clinical notes by conditioning transformer-based models on a limited window of past patient data.

In our work, instead of focusing on purely extractive or abstractive clinical summarization, we proposed an extractive-abstractive pipeline as a framework for clinical multi-document summarization. 
%The extractor is in charge of retrieving relevant information from the unstructured documents, and the abstractor is in charge of rewriting the extracted content into the format of the discharge summary section.

\paragraph{Faithfulness in Summarization.} 
%Measures that compare natrual languages, such as ROUGE~\citep{lin2003automatic} and BLEU~\citep{papineni2002bleu}, are often used to evaluate the quality of summarization by approximating the informativeness and fluency of the summary using an n-gram overlap-based approach. 
%However, as~\citet{schluter2017limits} and~\citet{cohan2016revisiting} argued, Rouge alone is insufficient and possibly misleading for measuring informativeness. 
Recognizing the limitation of the existing measures and the danger of hallucination in summarization systems, faithfulness in summarization has gained attention recently~\citep{kryscinski2020evaluating,cao2017faithful}. Recent work on faithfulness evaluation in summarization involves using textual entailment~\citep{maynez2020faithfulness} or question answer generation~\citep{arumae2019guiding,wang-etal-2020-asking}. For radiology summarization, \citet{zhang-etal-2020-optimizing} proposed using a radiology information extraction system to extract a pre-defined set of 14 pieces of factual information tailored to radiology reports.

In this paper, we approximate information overlap using the overlap of medical named entities. We argue that the domain of clinical encounter summarization is very different from the domains of most textual entailment tasks or question answer generation tasks. It is often much more specific, allowing us to apply the medical NER model. However, it is not as specific as the radiology summarization task, where a set of pre-defined information can more easily be identified.

\ignore{
\begin{enumerate}
    \item Physician Burnout, the fundamental cause and what can help alleviate the problem.
    \item Summarization Related: See et al., Chen \& Bansal, Liu \& Lapata, etc...
    \begin{enumerate}
        \item Extractive
        \item Abstractive
        \item Extract + Abstract
        \item Factualness in Summarization
    \end{enumerate}
    \item Clinical Summarization Related: generating SOAP (2020), summarizing radiology (2018, 2020), learning to write EHR (2018)
    \begin{enumerate}
        \item Extractive: the main focus of the conventional approaches.
        \item Abstractive: more recent, mostly on single radiology note
        \item Extract + Abstract: a clear trend in Clinical Summarization that leads to improve results both from factualness and rouge stand point
        \item How we differ: (to our best knowledge) first work focusing on generating discharge summary using multi-document summarization.
    \end{enumerate}
\end{enumerate}

Baselines

* Extractive baseline, something like this: https://arxiv.org/pdf/1809.04698.pdf
* Seq2Seq baseline, for instnce LSTM with attention: https://arxiv.org/abs/1409.0473
* Pointer Generator Networks: https://arxiv.org/abs/1704.04368
* Pre-trained seq2seq models like BART: https://arxiv.org/abs/1910.13461

* Learning to summarize radiology findings (https://arxiv.org/abs/1809.04698)
* Learning to write notes in EHR (https://arxiv.org/abs/1808.02622)
* Optimizing the Factual Correctness of a Summary: A Study of Summarizing Radiology Reports (https://arxiv.org/pdf/1911.02541.pdf)
* Generation and evaluation of artificial mental health records for Natural Language Processing (https://www.nature.com/articles/s41746-020-0267-x)
* Natural language generation for electronic health records (https://www.nature.com/articles/s41746-018-0070-0)
* Ontology-Aware Clinical Abstractive Summarization (https://arxiv.org/abs/1905.05818)
* Generating SOAP Notes from Doctor-Patient Conversations (https://arxiv.org/abs/2005.01795)

MIMIC III: https://mimic.physionet.org/gettingstarted/access/

*References*

* How to write a good discharge summary %(https://www.bmc.org/sites/default/files/Patient_Care/Primary_Care/Geriatrics/Education/Interns_and_Residents/field_Attachments/bmc-Transitions-of-Care.pdf)
}

% Talks about the two main family of models for summarizations.

% Extractive

% Abstractive

% Talks about extractive-abstractive pipeline, and how recent multi-document summarization work can be viewed in this framework (extractor is tf-idf, or logistic regression module). In clinical setting, you want to place more weight on the extractor.

% Talks about what is important in clinical summarization. Extractor is important. The work on abstractor. The danger of hallucination (well recognized)

% Existing evaluations of halluciation

\section{Dataset}
% \begin{table*}[ht]
%     \centering
%     \begin{adjustbox}{width=\textwidth}
%     \begin{tabular}{lrrrrrrr}
% \toprule
% {} &  brief hospital course &  history of present illness &  social history &  chief complaint &  medications on admission &  family history &  past medical history \\
% \midrule
% \# sent &              35.389029 &                   16.619268 &        4.929014 &         2.041833 &                  4.672851 &        2.630815 &              5.991150 \\
% \# word &             491.971417 &                  274.881153 &       44.903357 &         7.250368 &                 69.580294 &       17.026035 &             75.361568 \\
% \bottomrule
% \end{tabular}
% \end{adjustbox}
%     \caption{\hancomment{TODO: make text bigger, or make table horizontal?} Average number of words and sentences in the targeted medical sections}
%     \label{tab:stat_med_section}
% \end{table*}
\label{sec:dataset}

\begin{table}[t]
    \centering
    \begin{adjustbox}{width=\columnwidth}
    \begin{tabular}{lrrr}
    \toprule
    Dataset & Input & Ouput & \# Data \\
    \midrule
    Gigaword & $10^1$ & $10^1$ & $10^6$ \\
    CNN/DailyMail  & $10^2$–$10^3$ & $10^1$ & $10^5$ \\
    WikiSum  & $10^2$–$10^6$ & $10^1$–$10^3$ & $10^6$ \\
    Our Dataset & $10^4$–$10^5$ & *$10^0$–$10^3$ & $10^3$ \\
    \bottomrule
    \end{tabular}
    \end{adjustbox}
    \caption{Size comparison of summarization datasets. *For stats of the output sections, see Table~\ref{tab:ner_scores}.}
    \label{tab:prior_section}
    \vspace{-1em}
\end{table}

% \begin{table}[t]
%     \centering
%     \begin{tabular}{lrrr}
%     \toprule
%     \# doc & \# paragraph & \# sentences & \# words\\
%     \midrule
%     63.87 &  108.51 &  3527.73 & 36356.84\\
%     \bottomrule
%     \end{tabular}
%     \caption{Average number of documents, paragraphs, sentences, and words in prior clinical notes (source documents). For stats of the target sections, see Table~\ref{tab:ner_scores}.}
%     \label{tab:prior_section}
% \end{table}

%\hancomment{TODO: Add figures to explain the dataset (what is an encounter, what is a discharge summary) ?}

We derive our dataset from the MIMIC III database v1.4~\citep{johnson2016mimic}: a freely accessible, English-language, critical care database consisting of a set of de-identified, comprehensive clinical data of patients admitted to the Beth Israel Deaconess Medical Center's Intensive Care Unit (ICU). 
The database includes structured data such as medications and laboratory results and unstructured data such as clinical notes written by medical professionals. For this work, we will focus on the unstructured data.

The challenge for adapting the MIMIC III database for our purpose, however, is that MIMIC III is incomplete. Due to the way that MIMIC III was collected, not all clinical notes are available; only notes from ICU, radiology, echo, ECG, and discharge summary~\citep{johnson_shivade_2020} are guaranteed to be available. It is important to note that the incompleteness is not a property of the problem we are trying to address; it is a property of that database. We limit the incompleteness issue by focusing on the subset of encounters that contain at least one admission note (a clinical note written at the time of admission) as a proxy for completeness. This leaves us about 10\% of the total encounters, or around 6,000 encounters.

We identify seven medical sections in the discharge summary as our targets for summarization: (1) chief complaint, (2) family history, (3) social history, (4) medications on admission, (5) past medical history, (6) history of present illness, and (7) brief hospital course. \hancomment{TODO: add what is left out, and why} These medical sections were chosen based on their high prevalence in discharge summaries and their length diversity (see Table~\ref{tab:ner_scores}).

\paragraph{Target Section Extraction.} To extract the target medical sections from the discharge summary, we use a regular expression based approach to identify the medical section headers' variants from the training set. We then collect the content from the target medical section header and stop right before the next section header in the discharge summary. Around one hundred randomly selected extracted medical sections are manually examined to ensure no missing content or over-extraction. For each of these target medical sections, we then collect all the prior clinical notes (according to the chart date timestamp in MIMIC III) as their source documents. On average, the source documents consist of 64 documents and 36,3567 words. Table~\ref{tab:prior_section} shows a comparison with other dataset.%\footnote{Upon acceptance, we plan to work on releasing code that converts MIMIC III to our dataset.}

After the rule-based target extraction, we split the 6,000 encounters based on the \emph{subject id} to prevent data leakage. Each section is split into training, validation, and test set (80/10/10) using the same set of subject ids. If the rule-based target extraction returns nothing, the encounter is excluded. See Table~\ref{tab:ner_scores} for the statistic of sample size.

%By learning to compose a discharge summary from the prior documents found in the encounter, and by deriving the dataset from a public database, our work has the potential to be a benchmark for training and evaluation of encounter-level summarization.

\section{Models and Experiments}

\label{sec:model}

\ignore{

\begin{enumerate}
    \item Chen \& Bansal
    \begin{enumerate}
        \item Use as full model
        \item Use as an extractor
    \end{enumerate}
    \item Presumm EXT
    \item Pointer-Generator
    \item BART
\end{enumerate}
}

\hancomment{TODO: reformat this section more systematically. Plan (1) Talks about sentence-rewriting abstraction vs. long-paragraph abstraction, (2) Talks about RNN + RL(abs,ext), (3) Presumm(ext), (4) PointGen(abs), (5) BART(abs)}

\hancomment{TODO: Add figures to explain the models?}

\hancomment{TODO: Training details of the models. What we did to modify the models for our purpose. In the appendix?}

As explained in Section~\ref{sec:address}, our proposed pipeline involves an extractive summarization component and an abstractive summarization component. This section identifies a set of existing extractors and abstractors across a diverse range of different approaches to understand what models are suitable for encounter-level clinical summarization. To understand the robustness of these approaches, we train and test these models across seven medical sections with a diverse range of length.

\paragraph{\textbf{Extractors.}} Since our goal is to summarize an encounter conditioned on a targeted medical section, we focus our attention on supervised extractors. Supervised extractive summarization is often framed as a sentence extraction problem. Each sentence is encoded into a representation used to determine whether the sentence should be included in the extracted summary. RNN or transformer-based attention are often used to encode the surrounding sentences as context.
%for the extraction.

% \begin{enumerate}[leftmargin=1.2em]
%    \item 
\textbf{\textsc{RNN+RL}\textsubscript{ext}}: \newcite{chen2018fast} proposed a method to use reinforcement learning to fine-tune a pretrained RNN sentence extractor to a pointer network operating over sentences. By modeling the next sentence to extract (including the extra ``end-of-extraction'' sentence) as the action space, the current extracted sentences as the state space, and by using ROUGE between reference summary sentence and rewritten extracted sentence (rewritten by a separate pretrained abstractor) as the reward, the authors repurposed the sentence extractor to extract sentences from the source documents and reorder them as they might appear in the summary.
%    \item 

\textbf{\textsc{Presumm}\textsubscript{ext}}: \newcite{liu2019text} proposed Presumm, a family of summarization models. Here we are especially interested in the extractive summarization variant that uses a modified pretrained BERT model to encode sentences to determine whether the sentence should be included in the extracted summary. While the model has been shown to achieve competitive results, applying a BERT encoder to very long text can be challenging in terms of memory limitations. Thus, we apply a split-map-reduce framework, where the long text is split into smaller units during training and inference. After inference, each smaller unit's extracted sentences are then concatenated back together in the same order as appeared in the original source. Since the model only assigns scores to sentences, we sweep the score cutoff threshold on the validation set using ROUGE-L score, and apply that cutoff on the test set.
%\end{enumerate}

\paragraph{\textbf{Abstractors.}} 
%Abstractive summarization is often modeled as a sequence-to-sequence problem, mapping from source documents to the reference summary. 
In our extractive-abstractive pipeline, abstractors play a role in mapping the extracted sentences to the reference summary. Here we include two abstractor variants:\footnote{We also experimented with a pointer-generator~\cite{see2017get}, but we found that BART consistently outperforms pointer-generator, so we leave the results in the appendix.}
%The first is the sentence rewriter included in the Chen and Bansal work, which , and thus only have a local view when rewriting. The other two are commonly used abstractors, mapping from multiple sentences to multiple sentences, thus benefit from having a more global view.

%\begin{enumerate}[leftmargin=1.2em]
%    \item 
\textbf{\textsc{RNN+RL}\textsubscript{abs}}: Similar to \textsc{RNN+RL}\textsubscript{ext}, however, after each sentence is extracted, it is immediately rewritten by passing through a pretrained sentence-level abstractor. The goal is to rewrite each extracted sentence to the format of what might appear in the reference summary. This sentence-rewriting approach has the disadvantage of only having a local view when rewriting (thus no merging of information). However, the advantage is that the memory limitation of sentence-level rewriting does not grow with the number of sentences, so it can be applied to longer summaries. 
%In~\citet{chen2018fast}, the underlying abstractor is a Pointer-Generator~\cite{see2017get}.
%\item 

% \textbf{\textsc{PointGen}}: \newcite{see2017get} proposed a pointer-generator network combing (1) an encoder-decoder model with attention and (2) a copy mechanism that can \emph{point} back to the source documents and copy the content into the summary. The two mechanisms contribute differently depending on the source document and the generated content.
%\item 

\textbf{\textsc{BART}}: \newcite{lewis2019bart} propose \textsc{BART} as a transformer variant that uses a bidirectional encoder similar to BERT and an autoregressive (left to right) decoder similar to GPT. The model has competitive performance for summarization, and thus is our choice for transformer-based abstractor. In contrast to the sentence-rewriting approach of \textsc{RNN+RL}\textsubscript{abs}, we train \textsc{BART} to rewrite all the extracted sentences directly to the summary.
%\end{enumerate}

\paragraph{\textbf{Baselines.}}
Since clinical encounter summarization is a new task, there are no baselines from prior work. Following prior work on summarization, we include two special baselines: (1) \textsc{Oracle}\textsubscript{ext}: Extraction by using the reference summary; for each sentence in the reference summary, greedily select the source sentence in the source document that yields the maximum ROUGE-L score. (2) \textsc{Rule-based}\textsubscript{ext}: apply the same rule-based target section extraction method in in Section~\ref{sec:dataset} that was used to construct the dataset. Instead of applying to the discharge summary, we apply the same extraction method to the prior clinical documents.

\ignore{

\begin{enumerate}
    \item Oracle: cheating, using rouge L f1 score from summary sentences to extract source sentence.
    \item Rule-based: Apply the rule-based extraction module used to build the dataset to admission notes.
    \item Extractive Model Only
    \begin{enumerate}
        \item Chen \& Bansal EXT
        \item Presumm EXT
    \end{enumerate}
    \item Extractive + Abstractive
    \begin{enumerate}
        \item Chen \& Bansal E2E
        \item Chen \& Bansal EXT + BART
        \item Chen \& Bansal EXT + PointGen
        \item Presumm EXT + BART
        \item Presumm EXT + PointGen
    \end{enumerate}
\end{enumerate}
}

% To evaluate the effectiveness of our approach, we identify two families of use cases involving encounter-level clinical summarization. The first family corresponds to using an extractive summarization model, where the applications can range from highlighting, extraction in context, to chaining the extracted sentences as the summarization output. This family of use cases is desirable to clinical summarization, as extracted information can be linked back to their source, allowing inspection to support critical decision making. The second family corresponds to using an abstractive summarization model, where the application can range from an auto-complete aid (if the abstractive model is autoregressive) to smoothing and duplication removal for less critical sections. While this approach has the advantage of fluency, hallucination resulted from abstractive model is a real danger in the clinical use case, as should be properly measured.

\paragraph{Evaluating the extractor-abstractor pipeline.} For the two extractive models, \textsc{Rnn+RL}\textsubscript{ext} and  \textsc{PreSumm}\textsubscript{ext}, as well as the two extraction baselines, we report ROUGE scores as well as our proposed factualness-adjusted \{precision/recall/F\textsubscript{3}\} scores across the seven medical sections.

% -- we follow the convention and report Rouge scores. \hancomment{Ignore precision and recall description, will update shortly} Additionally, we also report medical NER-based precision and recall scores, where we treat the NER in the reference summary as ground-truth, and the NER in the extracted summary as the predictions. The precision in this case is a measure of over extraction, whereas the recall is a measure of coverage of the extraction model. We argue that instead of reporting f1, we should report them separately, as they convey two very different error modes.

For the abstractive models, we measure the combinations of abstractive models with extractive models in our proposed pipeline. This implies measuring the performance of three models (two pointer-generator models shown in Appendix): \textsc{Rnn+RL}\textsubscript{abs} (uses \textsc{Rnn+RL}\textsubscript{ext} as the extractor), \textsc{Rnn+RL}\textsubscript{ext} + \textsc{BART}, and \textsc{PreSumm}\textsubscript{ext} + \textsc{BART}. For the abstractors, we additionally measure \emph{incorrect hallucination rate} defined in Section~\ref{sec:measure}.

%\hancomment{Ignore precision and recall description, will update shortly} Besides reporting Rouge to measure fluency, we also report medical NER-based precision and recall scores. In this case, however, we additionally measure the \emph{information loss} and \emph{hallucination} attributed to going from the extracted output to the abstractive output. This is important as we need to correctly attribute the effect of using an abstractive system to weigh between the benefit and the danger. \hancomment{Ignore information loss and hallucination description, will update shortly} We measure the \emph{information loss} by measuring the name entities in the extracted output that is in the reference summary, but it is excluded after passing through the abstractive summarization system. For the \emph{hallucination} part, we measure novel name entities in the abstractive summary that is not in the extractive summary (regardless of the appearance of the name entities in the reference summary).

%To measure how hallucination manifest in different parts of the pipeline, we apply combinations of SOTA models. 

\vspace{-0.5em}

\section{Results and Discussion}

% \begin{table}[ht]
%     \centering
%     \begin{adjustbox}{width=\columnwidth}
%     \begin{tabular}{lrrrrr}
% \toprule
% {} &  \# words &  \# sents & RNN + RL\textsubscript{ext} &  Presumm\textsubscript{f1-ext} &  oracle \\
% medical sections           &        &        &                 &                 &         \\
% \midrule
% chief complaint            &   7.25 &   2.04 &  \textbf{44.95} &           11.94 &   72.89 \\
% family history             &  17.03 &   2.63 &  \textbf{39.96} &           32.87 &   55.27 \\
% social history             &  44.90 &   4.93 &  \textbf{36.65} &           35.43 &   60.96 \\
% medications on admission   &  69.58 &   4.67 &           42.08 &  \textbf{46.16} &   60.58 \\
% past medical history       &  75.36 &   5.99 &           46.33 &  \textbf{49.66} &   74.07 \\
% history of present illness & 274.88 &  16.62 &           33.41 &  \textbf{51.83} &   75.77 \\
% brief hospital course      & 491.97 &  35.39 &           18.63 &  \textbf{26.11} &   41.75 \\
% \bottomrule
% \end{tabular}
% \end{adjustbox}
%     \caption{Relationship between extractor performance vs. average word lengths of the medial sections. For shorter length section, RNN + RL\textsubscript{ext} outperforms Presumm\textsubscript{f1-ext}. Vise versa for longer length section.}
%     \label{tab:ext_length}
% \end{table}

\begin{figure*}[t]
\begin{subfigure}[t]{\columnwidth}
  \centering
  \includegraphics[width=\columnwidth]{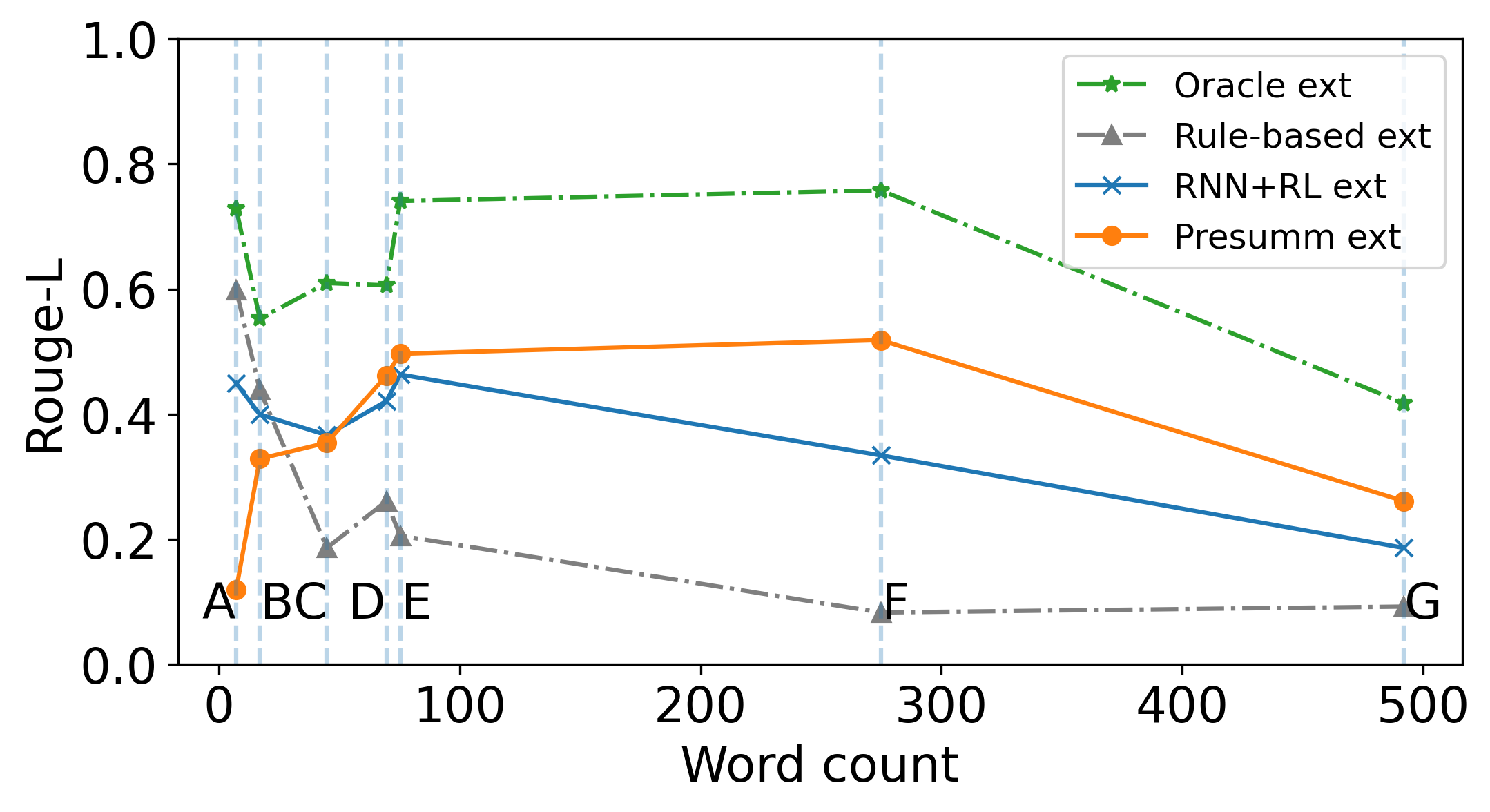}
  \caption{ROUGE-L of extractors vs. word count. 
  %For shorter length sections, RNN + RL\textsubscript{ext} outperforms Presumm\textsubscript{ext}. Vise versa for longer length sections. Maximum of the two extractive models shows a similar curve to the oracle extractive summary, with a constant drop.
  }
    \label{fig:rouge_scores_ext}
\end{subfigure}%
%\hspace{0.1\columnwidth}
\begin{subfigure}[t]{\columnwidth}
  \centering
  \includegraphics[width=\columnwidth]{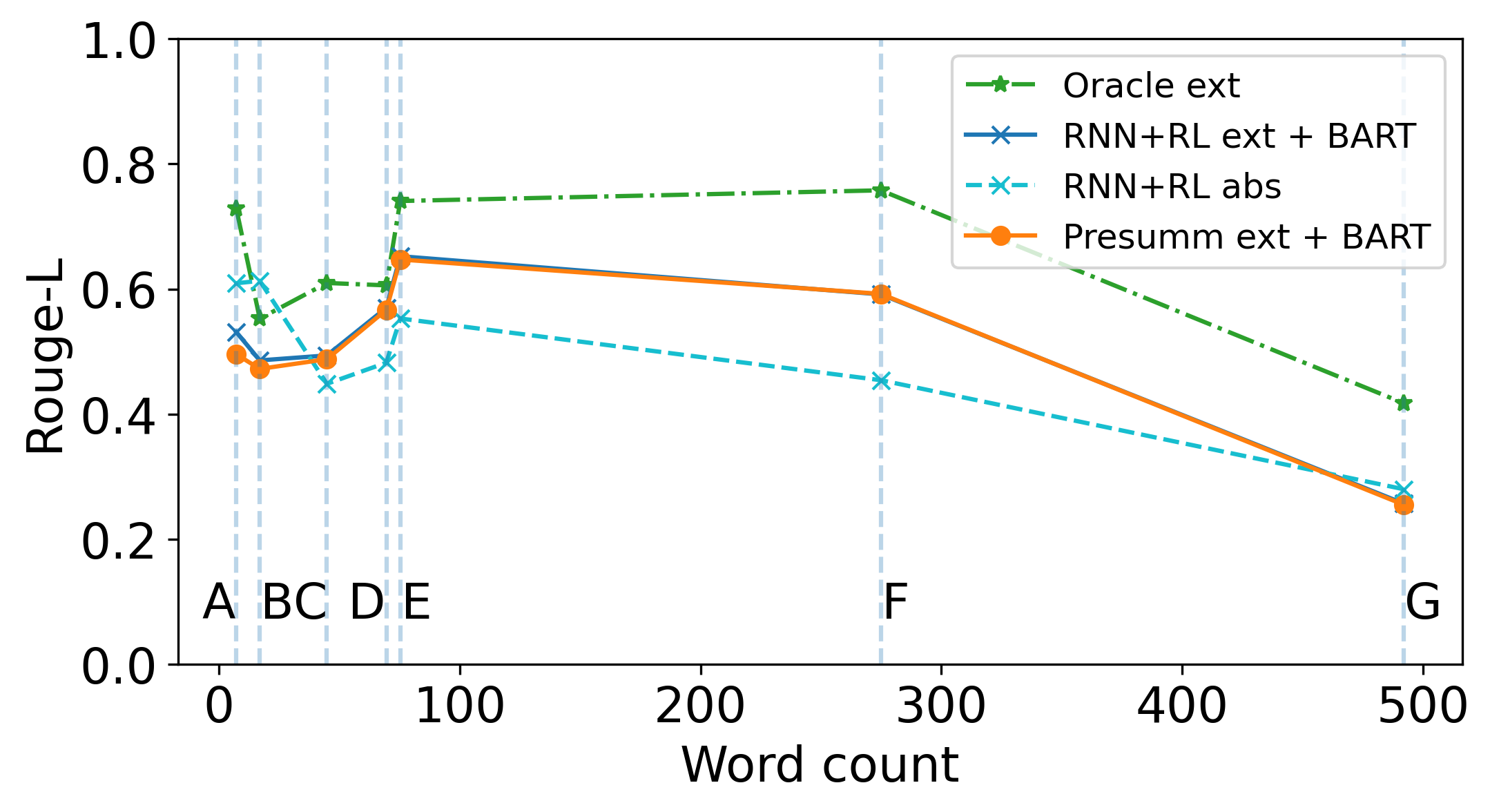}
  \caption{ROUGE-L of abstractors vs. word count. 
  %For shorter length sections, RNN + RL\textsubscript{abs} outperforms the two BART models. Vise versa for longer length sections. Interestingly, difference in Rouge-L of the two extractors disappear after passing through BART.
  }
  \label{fig:rouge_scores_abs}
\end{subfigure}
\caption{ROUGE-L of summarization models vs. average word lengths of the medical sections. Sections (dotted vertical lines) from short to long: (\textbf{A}) Chief complaint, (\textbf{B}) Family history, (\textbf{C}) Social history, (\textbf{D}) Medications on admission, (\textbf{E}) Past medical history, (\textbf{F}) History of present illness, and (\textbf{G}) Brief hospital course.}
\label{fig:rouge_scores_both}
\vspace{-0.5em}
\end{figure*}

% \begin{figure}[t]
%     \centering
%     \begin{adjustbox}{width=\columnwidth}
%     \includegraphics{extractor.png}
%     \end{adjustbox}
%     \caption{Relationship between extractor Rouge-L performance vs. average word lengths of the medial sections. For shorter length sections, RNN + RL\textsubscript{ext} outperforms Presumm\textsubscript{f1-ext}. Vise versa for longer length sections. Sections from short to long: (1) Chief complain, (2) Family history, (3) Social history, (4) Medications on admission, (5) Past medical history, (6) History of present illness, and (7) Brief hospital course}
%     \label{fig:rouge_scores_ext}
% \end{figure}

% \begin{figure}[t]
%     \centering
%     \begin{adjustbox}{width=\columnwidth}
%     \includegraphics{abstractor.png}
%     \end{adjustbox}
%     \caption{Relationship between abstractive Rouge-L performance vs. average word lengths of the medial sections. For shorter length sections, RNN + RL\textsubscript{abs} outperforms BART models. Vise versa for longer length sections. Interestingly, performance difference in the extractor does not contribute to the abstractive summaries produced by BART. The sections order is the same as Figure~\ref{fig:rouge_scores_ext}.}
%     \label{fig:rouge_scores_abs}
% \end{figure}

\begin{table*}[ht]
    \centering
    \begin{adjustbox}{width=\textwidth}
\begin{tabular}{llllllll}
\toprule
 & \makecell[l]{Chief Complaint} &  \makecell[l]{Family History} &  \makecell[l]{Social History} & \makecell[l]{Medications on \\ Admission} & \makecell[l]{Past medical \\ History} & \makecell[l]{History of \\ Present Illness} & \makecell[l]{Brief Hospital \\ Course} \\
\midrule
train / val / test & 4,757/559/625 & 4,686/555/614 & 4,677/552/618 & 4,689/557/616 & 4,746/558/623 & 4,754/559/625 & 4,758/558/625 \\
Output \# words            &         7.25 &          17.03 &         44.90 &                  69.58 &              75.36 &                    274.88 &               491.97 \\
Output \# sents            &         2.04 &         2.63 &         4.93 &                  4.67 &              5.99 &                    16.62 &                35.39 \\
\midrule
\midrule
\textsc{Oracle}\textsubscript{ext}         &  71.1/85.2/83.6 &  52.8/75.4/72.3 &  63.4/73.3/72.2 &           69.7/66.5/66.8 &       74.2/80.8/80.1 &             76.6/83.9/83.1 &        44.7/51.5/50.7 \\
\midrule
\textsc{Rule-based}\textsubscript{ext}     &  97.4/49.7/52.2 &  87.6/47.3/49.6 &  94.7/23.1/25.0 &           97.2/32.8/35.2 &       94.9/16.9/18.4 &               70.8/08.6/09.5 &           00.3/00.9/00.7 \\
\textsc{Presumm}\textsubscript{ext}        &  10.8/24.1/21.4 &  30.7/63.1/57.1 &  42.6/40.6/40.8 &           48.7/52.0/51.7 &       51.2/66.6/64.7 &             54.4/74.5/\textbf{71.9} &        26.5/47.7/44.2 \\
\textsc{RNN+RL}\textsubscript{ext}         &  44.2/72.8/\textbf{68.4} &  54.5/70.6/\textbf{68.6} &  43.2/71.0/\textbf{66.7} &           45.7/67.2/\textbf{64.2} &       43.6/81.7/\textbf{75.1} &             27.6/88.8/\textbf{72.7} &        15.3/69.7/\textbf{51.4} \\
\midrule
\textsc{Presumm}\textsubscript{ext} + \textsc{BART} &  45.5/63.6/61.2 &  46.1/70.2/66.7 &  60.0/66.0/65.3 &           67.1/77.7/76.5 &       69.7/73.3/72.9 &             68.0/64.5/64.8 &        37.4/26.8/27.6 \\
\textsc{RNN+RL}\textsubscript{ext} + \textsc{BART}  &  48.6/70.4/67.4 &  44.7/74.2/69.6 &  61.2/66.7/66.1 &           67.0/80.2/\textbf{78.7} &       70.0/74.6/\textbf{74.2} &             67.4/64.7/64.9 &        34.1/23.6/24.4 \\
\textsc{RNN+RL}\textsubscript{abs}         &  67.8/69.1/\textbf{69.0} &  75.8/73.0/\textbf{73.3} &  60.1/68.2/\textbf{67.3} &           70.9/69.0/69.2 &       64.7/68.8/68.3 &             40.8/82.2/\textbf{74.6} &        20.4/52.9/\textbf{45.6} \\

\bottomrule
\end{tabular}
\end{adjustbox}
    \caption{Dataset statistic and faithfulness-adjusted $\{Precision/Recall/F_{3}\}$ scores based on medical NER. 
    %displayed together with average number of words and sentences of each target medical sections.
    }
    \label{tab:ner_scores}
\vspace{-0.5em}
\end{table*}

\begin{figure}[t]
\centering
  \includegraphics[width=\columnwidth]{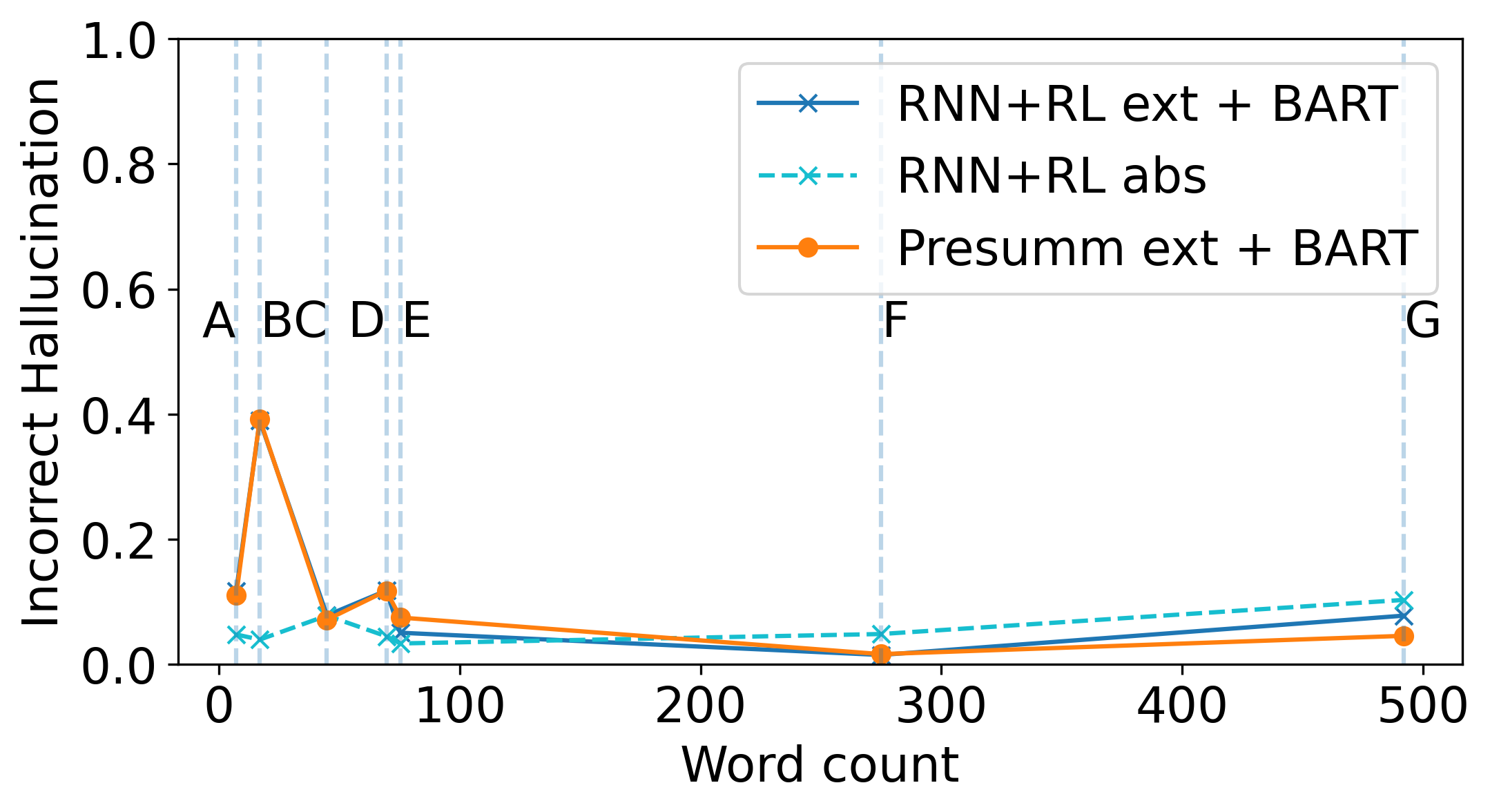}
\caption{NER-based incorrect hallucination rate of abstractive models vs. average word lengths. Extractors do not hallucinate. Order the same as Figure~\ref{fig:rouge_scores_both}.}
\label{fig:ner_hallucination}
\vspace{-0.5em}
\end{figure}

% \begin{figure*}[t]
% \begin{subfigure}[t]{0.95\columnwidth}
%   \centering
%   \includegraphics[width=0.95\columnwidth]{extractor_ner.png}
%   \caption{NER f1 score of extractor vs. word count.}
%     \label{fig:ner_scores_ext}
% \end{subfigure}%
% \hspace{0.1\columnwidth}
% \begin{subfigure}[t]{0.95\columnwidth}
%   \centering
%   \includegraphics[width=0.95\columnwidth]{abstractor_ner.png}
%   \caption{NER f1 score of abstractor vs. word count.}
%   \label{fig:ner_scores_abs}
% \end{subfigure}
% \caption{\hancomment{Ignore f1 score here, will update shortly} NER f1 of summarization models vs. average word lengths of the medial sections.  A similar pattern to Rouge-L is shown here. Sections order the same as~\ref{fig:rouge_scores_both}}
% \label{fig:ner_scores_both}
% \end{figure*}

\begin{table*}[ht]
    \centering
    \begin{adjustbox}{width=0.9\textwidth}
    \begin{tabular}{ll}
\toprule
{} &   Summary \\
\midrule
ground\_truth         &    \makecell[l]{past medical history : \# hypertension \# hyperlipidemia \# gerd \# ckd with baseline cr 1.3 \\ \# stable angina on long acting nitrate} \\ \midrule \midrule
Presumm\textsubscript{ext}           &  \makecell[l]{\# hypertension \# hyperlipidemia \# gerd \# ckd with baseline cr 1.3 nc occupation : changes \\ to medical and family history :} \\ \midrule
RNN+RL\textsubscript{ext}      &  \makecell[l]{\# simvastatin 20 mg once a day \# isosorbide mononitrate 40 mg once a day \# furosemide \\ 40 mg once a day \# pantoprazole 40 mg once a day  \# diltiazem xr 180 mg once a day \# tylenol \\ for gum pain \# proair hfa 90 mcg/actuation aerosol inhaler [ hospital1 ] prn \# prednisone per pt \\ 's son 2 weeks ago \# antibiotic for pneumonia per pt 's son 2 weeks ago past medical history : \# \\  hypertension \# hyperlipidemia \# gerd \# ckd with baseline cr 1.3 nc occupation : sinus rhythm .} \\ \midrule \midrule
Presumm\textsubscript{ext} + BART     &   \makecell[l]{past medical history : \# hypertension \# hyperlipidemia \# gerd \# ckd with baseline cr 1.3} \\ \midrule
RNN+RL\textsubscript{ext} + BART &    \makecell[l]{past medical history : \# hypertension \# hyperlipidemia \# gerd \# ckd with baseline cr 1.3} \\ \midrule
RNN+RL\textsubscript{abs} &  \makecell[l]{past medical history : \# hypertension \# hyperlipidemia \# gerd \# ckd with baseline cr 1.5 . .} \\
\bottomrule
\end{tabular}
\end{adjustbox}
    \caption{A random example showing summaries of past medical section. Despite RNN+RL\textsubscript{ext} over-extracted in this example, BART was able to smooth out the noise and generate the same output Presumm\textsubscript{ext} + BART. %Note that RNN+RL\textsubscript{abs} suffers from hallucination in this case (cr 1.3 -> cr 1.5).
    }
    \label{tab:qualitative_pmh}

\end{table*}

% \cshivadecomment{I feel moving the results table to the appendix may not be the best idea. Those are the **results** Figures 2a and 2b are more like discussion helping points. I understand the results table is huge and hence there is a space issue. My suggestion 1. Create a trimmed version of the table with points we want to drive home. (Pick one small and one large section, maybe?) Add it in the main paper 2. Keep the full table in appendix 3. Move some figures (currently in main paper) to appendix to make room for the trimmed table. This is my suggestion. But please talk to Philip and Doug and make a call.}

Extractive summarization and abstractive summarization are often applied in different settings and should thus be compared separately. For full results table, see Appendix A. Here we highlight the main findings. In Figure~\ref{fig:rouge_scores_ext}, we highlight the ROUGE-L scores (ROUGE-1 and ROUGE-2 have a similar pattern) of the two extractive summarization systems compared to the oracle and rule-based extractive summary. An interesting observation is the effect of length: average word count of the reference summary medical section. \textsc{RNN+RL}\textsubscript{ext} outperforms \textsc{Presumm}\textsubscript{ext} on shorter sections, and vice-versa for the longer sections. This difference can be partially attributed to the way \emph{cutoff} is being done at the extractors. For \textsc{RNN+RL}\textsubscript{ext}, an RL agent is trained to decide when to stop extracting sentences. For the shorter sections, the RL learns to stop at just a few sentences (e.g., a typical chief complaint has two sentences, family history has on average 2.6 sentences). On longer sections, however, we find that the RL agent has difficulty stopping, causing over-extraction. In contrast, for \textsc{Presumm}\textsubscript{ext}, a score cutoff threshold is tuned on the development set using the ROUGE-L score. 
%This difference in cutoff mechanisms mean that \textsc{RNN+RL}\textsubscript{ext} can easily stop at a small number of sentences (e.g., typical chief complaint has two sentences, family history has around 2.6 sentences). On the medium to longer secions, however, we found that \textsc{RNN+RL}\textsubscript{ext} tends to over-extract, hurting the ROUGE-L score. 
This approach has a more balanced performance but suffers at short sections. Another factor contributing to the lead of \textsc{Presumm}\textsubscript{ext} in the longer sections is our split-map-reduce framework, which enables the extractive model to conduct inference over all the clinical documents.

Interestingly, the baseline \textsc{Rule-based}\textsubscript{ext} performs surprisingly well on Rouge for the two shortest sections. Upon inspection, most of the extraction is just the medical section's title, without any content. This observation is backed up by the lower faithfulness-adjusted recall of this baseline.

%The NER-based f1 scores in Figure~\ref{fig:ner_scores_ext} show a similar story to ROUGE-L. Notably, even with the relatively high ROUGE-L scores, NER-based f1 scores suggest room for improvement specifically for an important application such as clinical summarization.

For abstractive summarization, we highlight the ROUGE-L of the three abstractors in Figure~\ref{fig:rouge_scores_abs}.
%BART consistently outperforms \textsc{PointGen} by a sizable margin. 
Interestingly, after being abstracted by BART, both \textsc{RNN+RL}\textsubscript{ext} and \textsc{Presumm}\textsubscript{ext} converged to roughly the same ROUGE-L scores. This suggests that in our extractor-abstractor pipeline, BART is effective in taking different extracted summaries and \emph{smoothing} them into the format and content expected for the medical sections. On the other hand, \textsc{RNN+RL}\textsubscript{abs} outperforms BART at the shorter sections, and even \textsc{Oracle}\textsubscript{ext} at the family history section. Note that \textsc{Oracle}\textsubscript{ext} is not necessarily an upper-bound for the abstractive summarization models; abstractors allow rewriting content in prior notes into the format of discharge summary. Sentence seqmentation (the basic unit of extraction) can also be noisy in clinical notes. On the other hand, the curve for \textsc{RNN+RL}\textsubscript{abs} is almost identical to \textsc{RNN+RL}\textsubscript{ext} in Figure~\ref{fig:rouge_scores_ext}, with a constant increase. This is largely attributed to the sentence-rewriting of the sentence-level abstractor that allows \textsc{RNN+RL}\textsubscript{abs} to keep the benefit of its extractor counterpart, while rewriting the content to reduce over-extracted sentences. 

%\hancomment{weak argument here, maybe need an example}

%\hancomment{will update shortly}

Table~\ref{tab:ner_scores} shows our faithfulness adjusted measures. For the extractors, \textsc{RNN+RL}\textsubscript{ext} out-performs all other extractors  on faithfulness-adjusted F\textsubscript{3} and even outperforms \textsc{Oracle}\textsubscript{ext} in the brief hospital course section. This is possible because \textsc{Oracle}\textsubscript{ext} is selected using ROUGE-L, not faithfulness-adjusted F\textsubscript{3}. For the abstractors, a similarly good performance is found for \textsc{RNN+RL}\textsubscript{abs}, where its precision consistently increases compared to \textsc{RNN+RL}\textsubscript{ext}. The good performance of \textsc{RNN+RL}\textsubscript{ext,abs} can largely be attributed to the high recall that has hurt their ROUGE-L performance in Figure~\ref{fig:rouge_scores_both}. Interestingly, the two BART models again perform roughly the same, with recall of \textsc{RNN+RL}\textsubscript{ext} + \textsc{BART} higher than \textsc{Presumm}\textsubscript{ext} + \textsc{BART}. For the longest section, generation for BART proves to be difficult, as indicated by the large drop of recall, whereas the sentence-wise rewriting strategy of \textsc{RNN+RL}\textsubscript{abs} has scaled better to longer sections.

The overall incorrect hallucination rate shown in  Figure~\ref{fig:ner_hallucination} is relatively low, with the notable exception of the family history section. Inspection of the generated summaries shows that the most common hallucination of both \textsc{BART} systems is the phrase \textit{``no family history''}. Interestingly, the ground truths corresponding to these hallucinations are mostly variations of the term \textit{``non-contributory''}; inspection of the source also shows that the family history section was often left blank. That being said, there are still cases of hallucinations where \textit{``no family history''} is followed by a condition (e.g., arrhythmia, cardiomyopathies) that is not mentioned in the source.

Table~\ref{tab:qualitative_pmh} shows a qualitative analysis of a randomly chosen summary of the past medical section. \hancomment{maybe need another example} In this case, RNN+RL\textsubscript{ext} over-extracted content from the previous sections. However, after passing through BART, BART successfully smooths out the noise and generates the same output as Presumm\textsubscript{ext} + BART. In this case, RNN+RL\textsubscript{abs} happens to be hallucinating (mapping cr 1.3 to cr 1.5). All summarization systems missed \textit{``\# stable angina on long acting nitrate''}; mention of \textit{``angina''} is actually not present in the prior clinical notes.

\vspace{-0.5em}

\section{Conclusion and Future Work}
\label{sec:conclusion}

\vspace{-0.5em}

We present a novel clinical summarization task -- discharge summary composition by summarizing prior clinical documents, derived from a public database (MIMIC III). By summarizing the vast number of clinical notes in a format clinicians are already trained to read and understand, our work has the potential to reduce the time clinicians spend on making sense of the data, allowing them to allocate more time to the patients.

We view this work as a promising first step to measure how existing models perform on the task and share the task with the community. One limitation of this work is that if there is novel information available only when writing the discharge summary, there will be no way of summarizing it. 
%We should therefore pay extra attention to hallucinations. 
It is also important to note that since we are using MIMIC III for training and evaluation, the results shown are biased to the dataset, as MIMIC III is an English-language collection from the ICU of a single hospital, and may not necessary be applicable to other clinical setting.

We identify three main challenges: (1) faithfulness, (2) evidence, and (3) long text. An extractor-abstractor pipeline is proposed to provide a natural way of fallback with an increasing amount of evidence at each fallback and also enable scaling to very long documents. To investigate the risk of hallucination and faithfulness in the summaries, we evaluate with a NER-based measure on top of ROUGE. Adapting state-of-the-art summarization models, our experiments over seven medical sections demonstrate the potential for the extractor-abstractor pipeline and represent a framework towards a set of enabling technologies that can assist clinicians to better make sense of the vast amount of unstructured data in the EHR.

% \subsection{Limitations}

% \hancomment{Move this forward?}

% 

\vspace{-0.5em}

\section{Ethical Considerations}

% \begin{itemize}
%     \item We use a public test collection
%     \item We do not identify any individuals
%     \item No IRB was needed
% \end{itemize}

\paragraph{Deidentification.} Our dataset is derived from the public database MIMIC III v1.4~\citep{johnson2016mimic}. \citet{johnson2016mimic} deidentified the database in accordance with the Health Insurance Portability and Accountability Act (HIPAA) standards. This standard requires removing all eighteen identifying data elements, including patient name, telephone number, address, and dates. These fields are replaced with placeholders. A constant (but different per patient) offset is applied to shift the dates. Patients over 89 years old were mapped to over 300, in compliance with HIPAA. 

Although under U.S. federal guidelines, secondary use of fully deidentified, publicly available data is exempt from institutional review board (IRB) review (45 CFR \S~46.104, ``Exempt research''), we still consider the dataset sensitive. We are careful to treat it as such. During training and error analysis, we of course do not attempt to identify individuals, and when the qualitative analysis is shown, we double-check to avoid showing potentially identifiable information.

\paragraph{Population.} In MIMIC III, out of the 38,161 patients, 71.34\% are White, 7.69\%  Black, 2.38\%  other, 2.37\%  Asian, and the rest unknown. Most of the patients in MIMIC III were older adults, with the most common age group being 71–80, followed by the 61–70 age group.~\citep{mimiciiistats}.

\paragraph{Broader Impact.} Clinical application has the genuine potential to affect people's lives. As we have emphasized in Section~\ref{sec:intro} and Section~\ref{sec:conclusion}, this work is not about a discussion for deployment, but rather a first step in understanding how the current existing summarization models perform. Importantly, we need to understand the failure modes of these systems and how to address these failures. Our emphasis on faithfulness and traceability of summarization reflects those beliefs. 
Hopefully, the three challenges we identify are the first of many future steps that will make progress toward alleviating the documentation burden of clinicians and ultimately result in a better quality of care for patients.

% % \section*{Acknowledgements}
\bibliography{anthology,custom}
\bibliographystyle{acl_natbib}

\appendix

\section{Appendix: Full Results}
\label{app:full_results}

\begin{table*}[ht]
    \centering
    \begin{adjustbox}{width=\textwidth}
    \begin{tabular}{llllllll}
\toprule
& \makecell[l]{Chief Complaint} &  \makecell[l]{Family History} &  \makecell[l]{Social History} & \makecell[l]{Medications on \\ Admission} & \makecell[l]{Past medical \\ History} & \makecell[l]{History of \\ Present Illness} & \makecell[l]{Brief Hospital \\ Course} \\
\midrule
Oracle ext             &  73.0/59.0/72.9 &  55.7/40.5/55.3 &  62.0/48.2/61.0 &           61.5/47.7/60.6 &       75.1/67.0/74.1 &             77.4/66.8/75.8 &        45.7/22.3/41.8 \\ \midrule
Rule-based ext         &  \textbf{59.8}/\textbf{44.5}/\textbf{59.8} &  \textbf{43.9}/\textbf{31.8}/\textbf{43.9} &  18.6/12.1/18.6 &           26.1/22.2/26.1 &       20.6/16.3/20.6 &                8.3/7.3/8.3 &           9.2/8.5/9.2 \\
RNN+RL ext             &  45.1/33.1/45.0 &  40.2/28.6/40.0 &  \textbf{37.6}/\textbf{27.2}/\textbf{36.6} &           43.4/35.6/42.1 &       47.9/40.2/46.3 &             34.8/28.3/33.4 &         21.3/6.7/18.6 \\
Presumm ext            &   12.3/6.9/11.9 &  33.2/24.0/32.9 &  36.3/\textbf{27.5}/35.4 &           \textbf{47.2}/\textbf{40.7}/\textbf{46.2} &       \textbf{50.8}/\textbf{41.9}/\textbf{49.7} &             \textbf{53.2}/\textbf{45.4}/\textbf{51.8} &        \textbf{29.6}/\textbf{10.6}/\textbf{26.1} \\
% Presumm ext F3         &   11.7/6.2/11.3 &  32.4/23.6/32.1 &  28.4/20.4/27.3 &           38.2/32.0/37.2 &       48.6/40.3/47.4 &             48.2/40.6/46.7 &         26.9/8.9/23.4 \\ 
\midrule
RNN+RL ext + PointGen  &  21.2/13.2/21.1 &  29.8/22.0/29.5 &  36.7/26.3/36.2 &           49.2/41.7/48.1 &       46.3/38.6/45.0 &             38.8/28.3/37.4 &         20.6/8.6/19.2 \\
Presumm ext + PointGen &  19.8/11.6/19.7 &  30.6/23.5/30.5 &  42.5/31.1/41.4 &           50.0/43.0/49.0 &       52.4/45.0/51.2 &             43.0/35.2/41.6 &         20.9/9.6/19.4 \\
RNN+RL ext + BART      &  53.5/37.5/53.1 &  48.9/38.6/48.6 &  \textbf{50.3}/\textbf{38.0}/\textbf{49.4} &           \textbf{58.2}/\textbf{51.9}/\textbf{57.0} &       \textbf{66.9}/\textbf{58.5}/\textbf{65.2} &             \textbf{61.1}/\textbf{51.3}/\textbf{59.1} &        28.2/10.6/25.7 \\
Presumm ext + BART     &  49.9/33.0/49.6 &  47.4/37.5/47.2 &  \textbf{49.6}/\textbf{38.3}/\textbf{48.8} &           \textbf{57.8}/50.9/56.7 &       \textbf{66.0}/\textbf{58.3}/\textbf{64.7} &             \textbf{61.0}/\textbf{52.4}/\textbf{59.2} &        28.0/\textbf{12.4}/25.5 \\
RNN+RL abs             &  \textbf{61.2}/\textbf{47.5}/\textbf{60.9} &  \textbf{61.6}/\textbf{50.5}/\textbf{61.3} &  45.9/33.7/44.8 &           49.9/42.2/48.2 &       57.5/47.9/55.3 &             47.6/38.4/45.4 &        \textbf{32.1}/10.4/\textbf{28.0} \\ \midrule
\# words                &         7.25037 &          17.026 &         44.9034 &                  69.5803 &              75.3616 &                    274.881 &               491.971 \\
\# sents                &         2.04183 &         2.63082 &         4.92901 &                  4.67285 &              5.99115 &                    16.6193 &                35.389 \\
\bottomrule
\end{tabular}
    \end{adjustbox}
    \caption{ROUGE-\{1/2/L\} scores, across different models and sections}
    \label{fig:rouge_scores}
\end{table*}

See Table~\ref{fig:rouge_scores} and Table~\ref{fig:factualness} for the full scores for all models on all seven sections.

\begin{table*}[ht!]
    \centering
    \begin{adjustbox}{width=\textwidth}
    \begin{tabular}{llllllll}
\toprule
 & \makecell[l]{Chief Complaint} &  \makecell[l]{Family History} &  \makecell[l]{Social History} & \makecell[l]{Medications on \\ Admission} & \makecell[l]{Past medical \\ History} & \makecell[l]{History of \\ Present Illness} & \makecell[l]{Brief Hospital \\ Course} \\
\midrule
\textsc{Oracle}\textsubscript{ext}         &  71.1/85.2/83.6 &  52.8/75.4/72.3 &  63.4/73.3/72.2 &           69.7/66.5/66.8 &       74.2/80.8/80.1 &             76.6/83.9/83.1 &        44.7/51.5/50.7 \\
\midrule
\textsc{Rule-based}\textsubscript{ext}     &  97.4/49.7/52.2 &  87.6/47.3/49.6 &  94.7/23.1/25.0 &           97.2/32.8/35.2 &       94.9/16.9/18.4 &               70.8/08.6/09.5 &           00.3/00.9/00.7 \\
\textsc{Presumm}\textsubscript{ext}        &  10.8/24.1/21.4 &  30.7/63.1/57.1 &  42.6/40.6/40.8 &           48.7/52.0/51.7 &       51.2/66.6/64.7 &             54.4/74.5/\textbf{71.9} &        26.5/47.7/44.2 \\
% \textsc{Presumm}\textsubscript{ext-F3}         &  10.2/25.7/22.3 &  29.5/64.8/57.9 &  25.6/48.0/44.1 &           34.1/57.3/53.6 &       47.7/71.0/67.7 &             47.0/78.7/73.7 &        19.5/67.9/54.4 \\
\textsc{RNN+RL}\textsubscript{ext}         &  44.2/72.8/\textbf{68.4} &  54.5/70.6/\textbf{68.6} &  43.2/71.0/\textbf{66.7} &           45.7/67.2/\textbf{64.2} &       43.6/81.7/\textbf{75.1} &             27.6/88.8/\textbf{72.7} &        15.3/69.7/\textbf{51.4} \\ \midrule
\textsc{RNN+RL}\textsubscript{ext} + \textsc{PointGen}  &  40.6/70.2/65.4 &  38.2/73.9/67.6 &  59.9/58.7/58.8 &           66.4/72.7/72.0 &       65.6/59.0/59.6 &             69.1/37.1/38.9 &        39.8/15.2/16.2 \\
\textsc{Presumm}\textsubscript{ext} + \textsc{PointGen} &  31.3/62.6/56.9 &  37.0/72.3/66.0 &  54.7/61.9/61.1 &           65.1/73.7/72.8 &       64.0/62.6/62.7 &             69.8/42.4/44.1 &        42.2/17.9/19.0 \\
\textsc{Presumm}\textsubscript{ext} + \textsc{BART} &  45.5/63.6/61.2 &  46.1/70.2/66.7 &  60.0/66.0/65.3 &           67.1/77.7/76.5 &       69.7/73.3/72.9 &             68.0/64.5/64.8 &        37.4/26.8/27.6 \\
\textsc{RNN+RL}\textsubscript{ext} + \textsc{BART}  &  48.6/70.4/67.4 &  44.7/74.2/69.6 &  61.2/66.7/66.1 &           67.0/80.2/\textbf{78.7} &       70.0/74.6/\textbf{74.2} &             67.4/64.7/64.9 &        34.1/23.6/24.4 \\
\textsc{RNN+RL}\textsubscript{abs}         &  67.8/69.1/\textbf{69.0} &  75.8/73.0/\textbf{73.3} &  60.1/68.2/\textbf{67.3} &           70.9/69.0/69.2 &       64.7/68.8/68.3 &             40.8/82.2/\textbf{74.6} &        20.4/52.9/\textbf{45.6} \\
\bottomrule
\end{tabular}
    \end{adjustbox}
    \caption{Faithfulness-adjusted $\{Precision/Recall/F_{3}\}$ scores based on medical NER.}
    \label{fig:factualness}
\end{table*}

\section{Appendix: Qualitative Analysis}
Table~\ref{tab:qualitative_soc} shows a random example on social history section.

\begin{table*}[th!]
    \centering
    \begin{adjustbox}{width=\textwidth}
    \begin{tabular}{ll}
\toprule
{} &   Summary \\
\midrule
ground\_truth         &    \makecell[l]{social history : retired from [ country 11150 ] . brother and son are part of support network .} \\ \midrule \midrule
Presumm\textsubscript{ext}           &  \makecell[l]{[ last name ( un ) 574 ] : retired gentleman from [ country \\ 4952 ] ; currently living with sons who are his main caretakers . . pt is hindi speaking only but able to \\ communicate his needs and pleasant and cooperative .} \\ \midrule
RNN+RL\textsubscript{ext}      &  \makecell[l]{family / social history : retired gentleman from [ country 4952 ] ; \\ currently living with sons who are his main caretakers . . saw pt ; did carotid massage ; give lopressor 50 mg \\ po bid starting tonight social history :} \\ \midrule \midrule
Presumm\textsubscript{ext} + BART     &   \makecell[l]{social history : [ last name ( un ) ] : retired gentleman \\ from [ country ] ; currently living with sons who are his main caretakers . pt is hindi speaking only but able to \\ communicate his needs and pleasant and cooperative .} \\ \midrule
RNN+RL\textsubscript{ext} + BART &    \makecell[l]{social history : retired gentleman from [ country 651 ] ; \\ currently living with sons who are his main caretakers .} \\ \midrule
RNN+RL\textsubscript{abs} &  \makecell[l]{social history : retired gentleman from [ country ] ] ; currently living \\ with sons who are his main caretakers .} \\
\bottomrule
\end{tabular}
\end{adjustbox}
    \caption{A random example showing summaries of social history section.}
    \label{tab:qualitative_soc}

\end{table*}

\section{Appendix: Reproducibility}
Here we describe the training details of the models for reproducibility.

\paragraph{\textsc{RNN+NL}\textsubscript{ext} and \textsc{RNN+NL}\textsubscript{abs}.}
Both models are trained following the original recipe~\cite{chen2018fast}. The training setup involves the following steps: (1) use gensim to train a word2vec embedding from scratch from the training set of the source documents, (2) construct pseudo pairs of sentences (source sentence, summary sentence): for each summary sentence, greedily finds the one-best source sentence using ROUGE-L recall, (3) use the pseudo pairs to train an RNN extractor, (4) use the pseudo pairs to train an pointer-generator that rewrites the sentences, and (5) train an RL agent that fine-tunes the RNN extractor with the sentence-rewriting pointer-generator. Model is trained on one V100 GPU, with an Adam optimizer of learning rate 1e-3. Here we use the same set of hyperparameters as~\citet{chen2018fast}. For more details, please refer to the original paper.

For each of the seven medical sections, we follow the training recipe, and repeat it five times. The reported models are chosen based on the validation set. We found that the RL fine-tuning step can potentially be very unstable. For the longer sections (e.g., brief hospital course and history of present illness), the RL fine-tuning can even fail to converge.

\paragraph{\textsc{Presumm}\textsubscript{ext}.}
We use the original implementation released with \textsc{Presumm}~\citep{liu2019text}. Learning rate is set to 2e-3 and extractor dropout rate is set to 0.1, following the original paper. \texttt{bert-base-uncased} is used as the pretrained BERT model. We made three important changes: (1) increase the maximum tokens the encoder can consume to 1024 tokens, (2) in the data preprocessing step, we construct pseudo pairs of sentences that will be later used to train the extractor: for each summary sentence, greedily finds the one-best source sentence using ROUGE-L recall, and (3) before the training begin, we split the source documents and their labels into segments smaller than 1024 tokens. After inference finishes, we concatenate the segments (together with a extraction score for each sentence) back together in the original order.

For each of the seven medical sections, we train the model on 4 V100 GPUs, with 150,000 training steps and model checkpointing every 2,000 steps. We report the model with the lowest model loss on the validation set. Since the model only assigns scores to sentences, we sweep the threshold of score cutoff on the validation set using ROUGE-L score, and apply that cutoff on the test set.

\paragraph{\textsc{PointGen}.}
We use an open implementation of pointer-generator~\citep{see2017get}, implemented with PyTorch and AllenNlp.\footnote{\url{https://github.com/kukrishna/pointer-generator-pytorch-allennlp}} Our model follows the original paper and has 256-dimensional hidden states and 128-dimensional word embeddings. The vocabulary size is set to 50k words for both source and target. The model is optimized using Adagrad with learning rate 0.15 and an initial accumulator value of 0.1, and trained on one v100 GPU for 50 epochs with early stopping on the validation set.

\paragraph{\textsc{BART}.}
We use the \texttt{Fairseq} \cite{ott2019fairseq} implementation of BART-large \cite{lewis2019bart} as it is shown to achieve the state-of-the-art ROUGE scores for abstractive summarization. We fine-tune the BART-large model with the standard learning rate of $3\times 10^{-5}$. We utilize a machine with 8 GPUs and batch size of 2048 input tokens per GPU. We train for a maximum of 10 epochs with early stopping to select the checkpoint with the smallest loss on the validation set. During decoding, we use beam search with beam size of 6. We restrict the generation length to be between 10 to 300 tokens.

\end{document}